\documentclass[conference]{IEEEtran}
\IEEEoverridecommandlockouts
\usepackage{cite}
\usepackage{amsmath,amssymb,amsfonts}
\usepackage{algorithmic}
\usepackage{graphicx}
\usepackage{textcomp}
\usepackage{xcolor}
\def\BibTeX{{\rm B\kern-.05em{\sc i\kern-.025em b}\kern-.08em
    T\kern-.1667em\lower.7ex\hbox{E}\kern-.125emX}}
\begin{document}

\title{Multi-Goal Optimal Route Planning Using the Cell Mapping Technique\\

}

\author{\IEEEauthorblockN{Athanasios Karagounis}
\IEEEauthorblockA{\textit{School of Science} \\
\textit{Department of Digital Industry Technologies}\\
GR34400, Psachna, Greece \\
akaragun@gs.uoa.gr}

}

\maketitle

\begin{abstract}
This manuscript explores the complexities of multi-objective path planning, aiming to optimize routes against a backdrop of conflicting performance criteria. The study integrates the cell mapping approach as its foundational concept. A two-pronged search strategy is introduced; initially, the cell mapping technique is utilized to develop a comprehensive database, encompassing all cells within the specified area. This database records the performance metrics for the most efficient routes from each cell to the designated target. The second phase involves analyzing this database to pinpoint the extent and count of all Pareto optimal routes from a selected starting cell to the target. This analysis contributes to solving the overarching multi-objective optimization challenge inherent in path planning. To validate this approach, case studies are included, and the results are benchmarked against the well-established multi-objective A* (MOA*) method. The study discovers that while the cell mapping method achieves similar outcomes to the MOA* method for routes originating from a single point, it demonstrates superior computational benefits, particularly when the starting and ending points are in separate, non-overlapping areas.
\end{abstract}


\section{Introduction}
Path planning is a crucial aspect of research in mobile robotics. This challenge is typically defined as the process of identifying feasible and safe routes from an initial point to a target location within a workspace. Concurrently, the identified paths must adhere to certain optimal criteria, such as minimal distance, reduced travel duration, and least energy consumption. Path planning must take into account two key elements: 1) the nature of the environment, which could be either static or dynamic, and 2) the robot's understanding of its surroundings. When a robot possesses complete environmental knowledge, the task is classified as global path planning. Conversely, when the robot's understanding of its surroundings is limited, the task falls under local path planning. This study introduces an effective approach to address the global multi-objective optimal path planning issue in static settings.\par
Numerous studies have been conducted to devise strategies for solving the global path planning problem \cite{goldman1994path}. These include various approaches like the roadmap method, the grid-based A* algorithm, hierarchical cell decomposition techniques, graph algorithms, and the artificial potential field method. In an effort to enhance the efficiency of these traditional methodologies, probabilistic algorithms have been introduced. Notable among these are the probabilistic roadmap method and the rapidly-exploring random trees. For a comprehensive understanding of both classical and probabilistic approaches, one can refer to specific studies. It's important to note that the majority of the existing path planning methods are primarily focused on addressing single-objective optimization challenges.\par
In practical scenarios, optimal path finding seldom involves a single objective. Real-world applications necessitate addressing multiple, and often conflicting, objectives at once. To address the multi-objective nature of path planning, several heuristic methods have been proposed \cite{amanatiadis2016multisensor}. The multi-objective extension of the A* algorithm, or MOA*, is discussed in certain studies \cite{stewart1991multiobjective}. The Firefly algorithm has also been examined. Other popular methodologies include Genetic Algorithms (GA) \cite{alajlan2013global} and Particle Swarm Optimization (PSO) \cite{kennedy1995particle}, which have been extensively explored. Additional methods that have garnered significant attention include neural networks, memetic algorithms, simulated annealing (SA) \cite{tavares2011simulated}, tabu search (TS), and cellular automata. Moreover, hybrid path planning algorithms have been developed, such as the integration of particle swarm optimization with the probabilistic roadmap method.\par

The cell-to-cell mapping technique, initially proposed by Hsu \cite{10.1115/1.3157686} for the comprehensive analysis of nonlinear dynamic systems, has recently shown promising results in uncovering global Pareto fronts with detailed structures for low and moderate dimensional multi-objective optimization problems (MOPs). The objective of our research is to investigate the capability of the cell mapping method in efficiently identifying all Pareto optimal paths within the context of multi-objective optimal path planning.\par
This study delves into the application of the cell mapping method to resolve multi-objective optimization problem (MOP) path planning challenges in static environments. It elaborates on the development of a cell space-based search algorithm, designed to pinpoint Pareto optimal paths from all cells to the goal cell. Additionally, the paper discusses the creation of a database system and a subsequent post-processing algorithm. These tools are instrumental in determining the coverage area of all Pareto optimal paths originating from a specific start cell to the target. The database also facilitates the extraction of the number of Pareto optimal paths along with their respective objective values. This process enables the construction of the Pareto front, a feature that is notably scarce in existing path planning literature.
\section{Multi-variate Optimal Path Planning}

The problem of multi-objective optimization in the context of global path planning is formulated as:
\begin{equation}
    \min_{k \in Q_{s,g}} F(k) = \min_{k \in Q_{s,g}} [f_1(k), \ldots, f_n(k)],
\end{equation}
where each \( f_i: \mathbb{R}^n \to \mathbb{R} \), with \( i = 1, 2, \ldots, n \), denotes an individual objective function that is subject to simultaneous minimization. The domain \( Q_{s,g} \) represents the set of all viable routes linking the start point \( s \) and the target \( g \). For this Multi-Objective Optimization Problem (MOP), optimal solutions are determined based on the principle of dominance, as outlined in subsequent definitions.
\begin{enumerate}
    \item[(a)] For paths \( i \) and \( j \in Q_{s,g} \), path \( j \) is said to be dominated by path \( i \) with respect to the MOP (1) if \( F(i) \preceq F(j) \) and \( F(i) \neq F(j) \). The relation \( F(i) \preceq F(j) \) signifies that every element of the vector \( F(i) \) is less than or equal to the corresponding element of \( F(j) \). If this is not the case, then \( j \) is considered non-dominated by \( i \).
    
    \item[(b)] A path \( i \in Q_{s,g} \) is termed Pareto optimal for the MOP (1) if there exists no path \( j \in Q_{s,g} \) that dominates \( i \).
    
    \item[(c)] The collection \( P \) of all Pareto optimal solutions is referred to as the Pareto set, denoted by

        $P := i \in Q_{s,g}$ $i$ is a Pareto optimal solution of the MOP (1).

    \item[(d)] The \( F(P) \) of \( P \) is known as the Pareto front.
\end{enumerate}

\section{Cell Mapping}

The cell mapping method discretizes the continuous point-wise space into a finite collection of discrete cells, each with integer coordinates. The total number of cells is finite, allowing them to be indexed by a single integer. The motion within a cell is represented by the movement of its central point, a concept known as the simple cell mapping \cite{koutras2021marsexplorer, chalvatzaras2022survey}.

To apply this method, divide the continuous 2D physical space into \( N = N_1 \times N_2 \) cells, assigning the objective vectors evaluated at the center of each cell. Finding a path involves identifying a sequence of cells that connects the start position cell to the target position cell. In a two-dimensional space, eight neighboring cells of any given cell are identified. 

In this work, we consider two objective functions or vectors, defined as:
\begin{equation}
    f_1(k) = \sum_{i,j \in P_k} \delta z(i, j); \quad
    f_2(k) = \sum_{i \in P_k} \delta \theta (i),
\end{equation}
where \( F(k) = [f_1(k), f_2(k)] \), \( P_k \) denotes the sequence of cells on the \( k \)th path, and \( i \) and \( j \) are neighboring cells in \( P_k \). The function \( \delta z(i, j) \) is given by:
\begin{equation}
    \delta z(i, j) = \sqrt{(x_i - x_j)^2 + (y_i - y_j)^2}.
\end{equation}
The function \( \delta \theta (i) \) indicates the cost associated with navigating through cell \( i \). The term \( \delta z(i, j) \) is used to quantify the direct distance between cell \( i \) and its adjacent cell \( j \), where \( (x_i, y_i) \) represent the coordinates of the center of cell \( i \). For simplicity in calculations, we define \( \delta z(i, j) \) based on the relative positioning of the cells: \( \delta z(i, j) = 10 \) when cell \( i \) and cell \( j \) lie along a horizontal or vertical axis, and \( \delta z(i, j) = 14 \) in cases where the cells are diagonally aligned.\par

The cell mapping technique facilitates the generation of a database encompassing all optimal paths and their corresponding objective values, which connect starting cells to the goal. This database, represented as \( B_k \), is constructed iteratively. The algorithm for database construction is outlined as follows:
\begin{equation}
    B_k := \{ [n_i, F_{n_i \to G}] \mid 1 \leq n_i \leq N \}, \tag{5}
\end{equation}
\begin{equation}
    F_{n_i \to G} := \{ F^r_{n_i \to G} \mid 1 \leq r \leq m(n_i) \},
\end{equation}
\begin{equation}
    F_{n_i \to G} \leftarrow \text{dom\_chk}(F_{n_i \to G}),
\end{equation}
Termination Condition: \( B_{k-1} = B_k \).

In this formulation, the database \( B_k \) stores information about free cells and their objective vectors leading to the goal cell, denoted as \( G \). The function \( m(n_i) \) denotes the count of distinct objective vectors \( F \) from cell \( n_i \) to the goal \( G \). It's important to note that the number of potential paths from cell \( n_i \) to \( G \) may differ from \( m(n_i) \).

In the database \( B_k \), \( n_i \) represents the index of cell \( n_i \). The superscript \( r \) in \( F^r_{n_i \to G} \) signifies the \( r \)th set of objective vectors \( F \) from cell \( n_i \) to \( G \). The objective vectors \( F^r_{n_i \to G} \) are calculated using the equation:
\begin{equation}
    F^r_{n_i \to G} = F^p_{n_j \to G} + \Delta F_{n_i \to n_j},
\end{equation}
where \( 1 \leq p \leq m(n_j) \). The incremental objective vectors \( \Delta F_{n_i \to n_j} \) are computed as per Equation (3).

The function \( \text{dom\_chk}(F_{n_i \to G}) \) refers to the dominance check operation applied to the objective vectors from cell \( n_i \) to \( G \). Through this operation, the non-dominated objective vectors are identified and selected. Subsequently, we elaborate on the steps involved in the cell space stratification mapping algorithm in a detailed manner.\par

The stratification mapping, incremental search \( \Delta F \), and dominance checking strategies are pivotal in ensuring that the database holds truly non-dominated objective vectors \( F_{n_i \to G} \) from every starting cell \( n_i \) to the goal \( G \). This database is comprehensive, containing information on multi-objective optimal paths from all initial conditions within the domain to the goal \( G \), which denotes the global nature of the proposed method.

Furthermore, the process of layer-wise updating and dominance checking results in Pareto optimal vectors that adhere to Bellman's principle of optimality. This principle asserts: "An optimal policy has the property that whatever the initial state and initial decision are, the remaining decisions must constitute an optimal policy with regard to the state resulting from the first decision" \cite{bellmann1957dynamic}. For instance, consider a non-dominated path \( P^* = (S, n_1, \ldots, n_i, n_{i+1}, \ldots, G) \) originating from the set \( S \) with objective vectors \( F_{P^*_{S \to G}} \subseteq B_k \), and its subpath \( P^*_i = (n_i, n_{i+1}, \ldots, G) \) with objective vectors \( F_{P^*_{i_{n_i \to G}}} \subseteq B_k \). The algorithm confirms that \( F_{P^*_{i_{n_i \to G}}} \subseteq F_{n_i \to G} \) and \( P^*_i \) continue to be non-dominated and hence, remain Pareto optimal.

Additionally, the computational complexity of the cell mapping algorithm does not vary with the number of cells \( N_G \) in \( G \). It's also noteworthy that while all optimal paths from the set \( S \) are likely to converge at the goal \( G \), they may not encompass the entire set comprehensively.

\section{Experimental Results}

We illustrate the application of the cell mapping method in identifying Pareto optimal paths and constructing the Pareto front for specified start and goal cells. The outcomes of the Pareto front and the subsequent post-processing utilizing the cell mapping method are presented.

A space is segmented into \( 117 \times 117 \) cells. We designate a cell as a starting cell, and a cell as the goal. The height of the terrain is considered as the terrain cost. The Pareto set, representing the coverage area of optimal paths from the start cell to the goal using the cell mapping method with the Pareto front for optimal paths determined by the cell mapping algorithm. This approach yields 255 non-dominated objective vectors (forming the Pareto front) and 15089 optimal paths. Notably, the Pareto front demonstrates a locally non-convex characteristic, underscoring the complexity inherent in the underlying multi-objective optimal path planning problem.\par

In the forthcoming examples, we aim to evaluate the computational efficiency and the accuracy of the solutions derived through the cell mapping method, in relation to the MOA* method. A blue cell is designated as the goal \( G \). The terrain cost \( \delta \theta \) is set to 1 for magenta cells and 0 for others. Through this straightforward example, the global solution capability of the cell mapping method is demonstrated.

The directions of optimal paths from all cells to \( G \), as encoded in the database \( B_k \). Consequently, the database can furnish optimal paths from any initial cell to the goal. For instance, if we select the green cell as the start cell \( S \), it is feasible to extract all Pareto optimal paths connecting \( S \) to \( G \). It is important to note that an identical set of Pareto optimal paths from the start cell \( S \) to the goal \( G \) could be obtained using the MOA* algorithm within a comparable computational timeframe. The solution  represents a subset. To achieve the same global solution with the MOA* algorithm, it would be necessary to run the algorithm \( N - 1 \) times, where \( N \) denotes the total number of cells in the domain.

Our examination of the computational efficiency of the cell mapping method, alongside a comparison of solution accuracy with the MOA* method \cite{mandow2005new}, involved 100 complex random maps with varying divisions of the domain. We discovered that both methods yield identical sets of Pareto optimal paths for a given start cell to the same goal. The CPU time required by the cell mapping method to create a database representing global solutions is approximately equivalent to that needed by the MOA* algorithm to generate Pareto optimal paths for one start cell, and can be run in edge devices \cite{faniadis2020deep}.

It is also noteworthy that in scenarios where Pareto optimal paths are needed from a multitude of starting locations converging to a common target, the cell mapping method exhibits its full potential. This is particularly true for applications requiring multi-objective path planning to connect one or more disjoint areas. In such cases, the cell mapping method provides a substantial advantage over the MOA* algorithm and other methods. Next, we will present a map to plan multi-optimal paths connecting disjoint goal areas.

\section{Conclusions}
This paper introduced an efficient approach using the cell mapping method for addressing multi-objective optimal path planning challenges. Central to the study were objectives like minimizing path length and reducing terrain cost. A comprehensive global database, containing the optimal objective vectors (Pareto front) for routes from all domain cells to the target, was developed employing this method. Additionally, the extent of the Pareto optimal paths (Pareto set) and the total count of all feasible paths from a selected start cell \( S \) to the destination \( G \) were derived through a specialized post-processing algorithm applied to the database. The efficacy of this innovative method was underscored through comparative analyses with the established MOA* method.

\bibliographystyle{IEEEtran}
\bibliography{arxivbibtex}

\end{document}